\pdfoutput=1
\newcommand{\AgentPromptRoundOne}{Answer the following multiple choice question. The last line of your response should be of the following format: 'Answer: \$LETTER' (without quotes) where LETTER is one of ABCD. Think step by step before answering.\newline \newline Which of these combinations of clinical features is most suggestive of mixed mitral valve disease with a predominance of mitral regurgitation?\newline \textcolor{green}{A. Displaced apex beat; soft first heart sound; pan-systolic murmur; short mid-diastolic murmur}\newline B. Irregularly, irregular pulse; displaced apex beat; ejection systolic murmur; short mid-diastolic murmur\newline C. Displaced apex beat; normal first heart sound; pan-systolic murmur; long mid-diastolic murmur\newline D. Tapping apex beat; loud first heart sound; pan-systolic murmur; long mid-diastolic murmur
}
\newcommand{\AgentOneResponseRoundOne}{
Based on the given clinical features, option C: Displaced apex beat; normal first heart sound; pan-systolic murmur; long mid-diastolic murmur is least suggestive of mixed mitral valve disease with a predominance of mitral regurgitation.\newline \newline The features most suggestive of mixed mitral valve disease with a predominance of mitral regurgitation are:\newline \newline 1. Displaced apex beat: This can occur due to chronic volume overload or dilatation of the left ventricle from mitral regurgitation.\newline 2. Soft or absent first heart sound: This can be due to the decreased force of mitral valve closure during diastole.\newline 3. Pan-systolic murmur: This is typically heard in mitral regurgitation due to the continuous regurgitant jet throughout the cardiac cycle.\newline 4. Long mid-diastolic murmur: This can occur due to the continued mitral regurgitant jet during diastole.\newline \newline Therefore, the correct answer is:\newline \newline A. Displaced apex beat; soft first heart sound; pan-systolic murmur; long mid-diastolic murmur\newline \newline \textcolor{green}{Answer: A}
}
\newcommand{\AgentTwoResponseRoundOne}{
Based on the given clinical features, the combination most suggestive of mixed mitral valve disease with a predominance of mitral regurgitation would be:\newline \newline C. Displaced apex beat; normal first heart sound; pan-systolic murmor; long mid-diastolic murmur\newline \newline Explanation:\newline The presence of a displaced apex beat and a pan-systolic murmur points towards mitral valve disease. However, the duration and timing of the additional diastolic murmur helps distinguish the underlying etiology. A long mid-diastolic murmur is more suggestive of functional mitral regurgitation, which is a component of mixed mitral valve disease with a predominance of mitral regurgitation. Therefore, option C is the correct answer.\newline \newline \textcolor{red}{Answer: C}
}
\newcommand{\AgentThreeResponseRoundOne}{
The combination of a displaced apex beat, a pan-systolic murmor, and a long mid-diastolic murmur is most suggestive of mitral stenosis. Thus, the answer to the question is D. however, we were asked for a combination that is most suggestive of mixed mitral valve disease with a predominance of mitral regurgitation. With this in mind, the correct answer is: C. Displaced apex beat; normal first heart sound; pan-systolic murmur; long mid-diastolic murmur\newline \newline Explanation: Mixed mitral valve disease refers to a condition in which both mitral regurgitation and mitral stenosis are present. The presence of a predominantly regurgitant jet suggests a more advanced state of the disease. \newline A displaced apex beat can be seen in conditions of volume overload, such as mitral regurgitation, and is less common in mitral stenosis. A normal first heart sound and a long mid-diastolic murmur are more likely to be heard in mitral regurgitation than mitral stenosis, as they represent the opening and closing of the mitral valve during diastole. Therefore, option C, with a displaced apex beat, normal first heart sound, and a long mid-diastolic murmur, is the most suggestive of mixed mitral valve disease with a predominance of mitral regurgitation.\newline \newline \textcolor{red}{Answer: C}.
}

\newcommand{\AgentOnePromptRoundTwo}{
These are solutions and confidence values from 1 to 10 (higher means more confident) to the problem from other agents: \newline \newline  One agent solution (confidence level is 1): ```\textcolor{red}{<agent 2 response>}```\newline \newline  One agent solution (confidence level is 1): ```\textcolor{red}{<agent 3 response>}```\newline \newline Based off the opinion of other agents, can you give an updated response? Do not mention your confidence. Think step by step before answering. The last line of your response should be of the following format: 'Answer: \$LETTER' (without quotes) where LETTER is one of ABCD.
}
\newcommand{\AgentTwoPromptRoundTwo}{
These are solutions and confidence values from 1 to 10 (higher means more confident) to the problem from other agents: \newline \newline  One agent solution (confidence level is 10): ```\textcolor{green}{<agent 1 response>}```\newline \newline  One agent solution (confidence level is 1): ```\textcolor{red}{<agent 3 response>}```\newline \newline Based off the opinion of other agents, can you give an updated response? Do not mention your confidence. Think step by step before answering. The last line of your response should be of the following format: 'Answer: \$LETTER' (without quotes) where LETTER is one of ABCD.
}
\newcommand{\AgentThreePromptRoundTwo}{
These are solutions and confidence values from 1 to 10 (higher means more confident) to the problem from other agents: \newline \newline  One agent solution (confidence level is 10): ```\textcolor{green}{<agent 1 response>}```\newline \newline  One agent solution (confidence level is 1): ```\textcolor{red}{<agent 2 response>}```\newline \newline Based off the opinion of other agents, can you give an updated response? Do not mention your confidence. Think step by step before answering. The last line of your response should be of the following format: 'Answer: \$LETTER' (without quotes) where LETTER is one of ABCD.
}

\newcommand{\AgentOneResponseRoundTwo}{
Based on the information provided by the other agents, the combination most suggestive of mixed mitral valve disease with a predominance of mitral regurgitation is:\newline \newline A. Displaced apex beat; soft first heart sound; pan-systolic murmur; long mid-diastolic murmur.\newline \newline The presence of a displaced apex beat, soft first heart sound, and prolonged pan-systolic murmur indicate mitral valve disease. The duration and timing of the diastolic murmur can be used to distinguish between mitral stenosis and mitral regurgitation. A long, holodiastolic murmur is more commonly heard in mitral regurgitation than mitral stenosis. Therefore, the presence of a long mid-diastolic murmur with a displaced apex beat and soft first heart sound suggests mixed mitral valve disease with a predominance of mitral regurgitation.\newline \newline \textcolor{green}{Answer: A}.
}
\newcommand{\AgentTwoResponseRoundTwo}{
Based on the clinical features provided and the consensus of other agents, option C: Displaced apex beat; normal first heart sound; pan-systolic murmur; long mid-diastolic murmur is the most suggestive of mixed mitral valve disease with a predominance of mitral regurgitation.\newline \newline \textcolor{red}{Answer: C}.
}
\newcommand{\AgentThreeResponseRoundTwo}{
Based on the discussions among different agents, the most suggestive combination of clinical features for mixed mitral valve disease with a predominance of mitral regurgitation is:\newline \newline A. Displaced apex beat; soft first heart sound; pan-systolic murmur; long mid-diastolic murmur\newline \newline Explanation:\newline Mixed mitral valve disease is characterized by the co-existence of mitral stenosis and mitral regurgitation. The presence of a displaced apex beat, soft first heart sound, pan-systolic murmur, and long mid-diastolic murmur can suggest a diagnosis of mixed mitral valve disease with a predominance of mitral regurgitation. The displaced apex beat can occur due to chronic volume overload or dilatation of the left ventricle, as seen in mitral regurgitation. The soft or absent first heart sound can be due to the decreased force of mitral valve closure in mitral regurgitation. The pan-systolic murmur is typically heard in mitral regurgitation due to the continuous regurgitant jet throughout the cardiac cycle. The long mid-diastolic murmur can occur due to the continued mitral regurgitant jet during diastole. Therefore, the combination of these clinical findings, A. Displaced apex beat; soft first heart sound; pan-systolic murmur; long mid-diastolic murmur, is most suggestive of mixed mitral valve disease with a predominance of mitral regurgitation.\newline \newline \textcolor{green}{Answer: A}.
}

\newcommand{\AgentOnePromptRoundThree}{
These are solutions and confidence values from 1 to 10 (higher means more confident) to the problem from other agents: \newline \newline  One agent solution (confidence level is 1): ```\textcolor{red}{<agent 2 response>}```\newline \newline  One agent solution (confidence level is 10): ```\textcolor{green}{<agent 3 response>}```\newline \newline Based off the opinion of other agents, can you give an updated response? Do not mention your confidence. Think step by step before answering. The last line of your response should be of the following format: 'Answer: \$LETTER' (without quotes) where LETTER is one of ABCD.
}
\newcommand{\AgentTwoPromptRoundThree}{
These are solutions and confidence values from 1 to 10 (higher means more confident) to the problem from other agents: \newline \newline  One agent solution (confidence level is 10): ```\textcolor{green}{<agent 1 response>}```\newline \newline  One agent solution (confidence level is 10): ```\textcolor{green}{<agent 3 response>}```\newline \newline Based off the opinion of other agents, can you give an updated response? Do not mention your confidence. Think step by step before answering. The last line of your response should be of the following format: 'Answer: \$LETTER' (without quotes) where LETTER is one of ABCD.
}
\newcommand{\AgentThreePromptRoundThree}{
These are solutions and confidence values from 1 to 10 (higher means more confident) to the problem from other agents: \newline \newline  One agent solution (confidence level is 10): ```\textcolor{green}{<agent 1 response>}```\newline \newline  One agent solution (confidence level is 1): ```\textcolor{red}{<agent 2 response>}```\newline \newline Based off the opinion of other agents, can you give an updated response? Do not mention your confidence. Think step by step before answering. The last line of your response should be of the following format: 'Answer: \$LETTER' (without quotes) where LETTER is one of ABCD.
}

\newcommand{\AgentOneResponseRoundThree}{
Based on the discussions among different agents, the combination of clinical features most suggestive of mixed mitral valve disease with a predominance of mitral regurgitation is:\newline \newline A. Displaced apex beat; soft first heart sound; pan-systolic murmur; long mid-diastolic murmur.\newline \newline Explanation:\newline Mixed mitral valve disease is defined as the presence of both mitral stenosis and mitral regurgitation. The clinical presentation of mixed mitral valve disease can vary depending on the severity and predominance of each component.\newline \newline A displaced apex beat, soft first heart sound, pan-systolic murmur, and long mid-diastolic murmur are more indicative of mitral regurgitation than mitral stenosis. The displaced apex beat can result from the increased volume load in the left ventricle due to mitral regurgitation. A soft or absent first heart sound may be heard in mitral regurgitation due to the decreased force of mitral valve closure, while the pan-systolic murmur and long mid-diastolic murmur are characteristic features of mitral regurgitation.\newline \newline Therefore, the combination of these clinical findings, A. Displaced apex beat; soft first heart sound; pan-systolic murmur; long mid-diastolic murmur, is most suggestive of mixed mitral valve disease with a predominance of mitral regurgitation.\newline \newline \textcolor{green}{Answer: A}.
}
\newcommand{\AgentTwoResponseRoundThree}{
Based on the consensus of other agents, the combination most suggestive of mixed mitral valve disease with a predominance of mitral regurgitation is: A. Displaced apex beat; soft first heart sound; pan-systolic murmur; long mid-diastolic murmur.\newline \newline \textcolor{green}{Answer: A}.
}
\newcommand{\AgentThreeResponseRoundThree}{
Based on the information provided by various agents and the consensus reached, the combination of clinical features most suggestive of mixed mitral valve disease with a predominance of mitral regurgitation is:\newline \newline A. Displaced apex beat; soft first heart sound; pan-systolic murmur; long mid-diastolic murmur\newline \newline Here's the explanation:\newline The presence of a displaced apex beat and a pan-systolic murmur suggest mitral valve disease. However, the duration and timing of the diastolic murmur can help distinguish between mitral stenosis and mitral regurgitation. Mitral regurgitation is more commonly associated with a long, holodiastolic or mid-diastolic murmur, while mitral stenosis typically has a short, mid-diastolic murmur. In this scenario, the long mid-diastolic murmur, along with the displaced apex beat and soft first heart sound, makes mixed mitral valve disease with a predominance of mitral regurgitation the most likely diagnosis. Therefore, the answer is:\newline \newline \textcolor{green}{Answer: A}.
}

\documentclass[11pt]{article}
\usepackage{multirow}
\usepackage{tabu}

\usepackage{xltabular}
\usepackage{array}      % For advanced column formatting
\usepackage{booktabs}   % For nicer horizontal lines

\usepackage[preprint]{acl}
\usepackage{nameref}

\usepackage{times}
\usepackage{latexsym}
\usepackage{amsmath}
\usepackage{hyperref}
\usepackage{tcolorbox}
\usepackage{enumitem}

\usepackage{float}
\usepackage{color,soul}

\usepackage[T1]{fontenc}
\usepackage[utf8]{inputenc}

\usepackage{microtype}

\usepackage{inconsolata}

\usepackage{graphicx}

\title{DebUnc: Improving Large Language Model Agent Communication With Uncertainty Metrics}

\author{Luke Yoffe \qquad Alfonso Amayuelas \qquad William Yang Wang \\
        University of California, Santa Barbara \\
        \texttt{\{lukeyoffe, amayuelas\}@ucsb.edu, william@cs.ucsb.edu}
}

\begin{document}
\maketitle
\begin{abstract}
Multi-agent debates have been introduced to improve the accuracy of Large Language Models (LLMs) by having multiple agents discuss solutions to a problem over several rounds of debate. However, models often generate incorrect yet confident-sounding responses, which can mislead others. This issue arises partly because agents do not consider how confident their peers are. To address this, we propose DebUnc, a debate framework that uses uncertainty metrics to assess agent confidence. Confidence is then conveyed through a modified attention mechanism that adjusts token weights, or through textual prompts. Evaluations across benchmarks show that attention-based methods are particularly effective and that performance continues to improve as uncertainty estimation becomes more reliable. The code is available at \url{https://github.com/lukeyoffe/debunc}.

\end{abstract}

\section{Introduction}
Large language models (LLMs) have demonstrated impressive performance across various domains, including law, academia, and coding \citep{gpt4}. However, they are prone to hallucinations, where they confidently generate false or misleading information \citep{survey_hallucination}. This poses significant risks in real-world applications. For example, an LLM tutor providing incorrect explanations could mislead students, while a customer service agent giving faulty advice could frustrate users. In high-stakes fields like healthcare, journalism, or finance, hallucinations can have severe consequences, including financial loss or health risks.  To mitigate these issues, researchers have explored multi-agent debate, where multiple LLMs propose diverse solutions and critique each other’s reasoning over several rounds of debate\citep{divergent_thinking}.This process has been shown to enhance the reasoning and accuracy of LLMs, outperforming simpler approaches such as majority voting or chain-of-thought prompting \citep{multiagentdebate}.

\definecolor{yellow}{RGB}{255, 171, 64}
\definecolor{red}{RGB}{207, 0, 0}
\definecolor{green}{RGB}{57, 185, 0}
\begin{figure}[ht]
    \centering
    \includegraphics[width=\columnwidth]{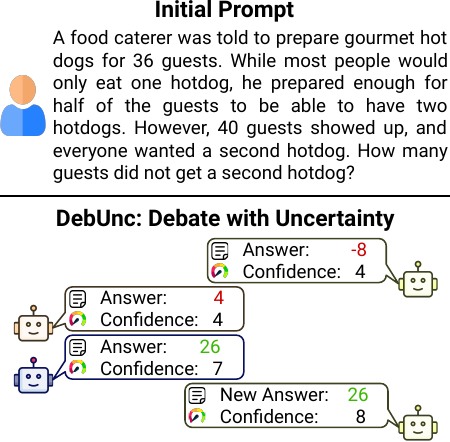}
\caption{Example three-agent debate. The first agent initially provides an incorrect response but corrects itself after considering the answers and confidence levels of others. Each agent uses a LLM to generate its response and an uncertainty metric to assesses its confidence. Correct answers are shown in \textcolor{green}{green}, while incorrect ones are shown in \textcolor{red}{red}.}
\vspace{-5mm}
    \label{fig:chat}
\end{figure}

Ideally, debate should help agents recognize and correct errors when some provide incorrect answers. In practice, while agents often agree on the same final answer, it is sometimes incorrect. This issue stems from flawed communication, as LLMs typically respond with high confidence regardless of accuracy \citep{multiagentdebate}. A confidently incorrect response can mislead other agents, causing the system to converge on an incorrect conclusion. Because it is difficult to gauge a LLM's certainty based on its response alone, uncertainty metrics have been developed to provide a more reliable confidence measure.

By contrast, people often use qualifiers such as "I am sure that..." or "I am not sure, but..." to express confidence during discussions. These cues help others assess the reliability of information.

Building on these insights, we present DebUnc, a novel multi-agent debate framework that combines multi-agent \textbf{Deb}ates and model \textbf{Unc}ertainty metrics. After each round of debate, we measure each agent's confidence with an uncertainty metric. In the following round, both the agents' responses and confidences are shared with the other agents. We explore two methods for communicating agent uncertainty: incorporating the uncertainty directly into the textual prompt, as shown in \autoref{fig:pipeline}, and shifting the model's attention towards more confident agents, as depicted in \autoref{fig:attn}. We evaluate DebUnc across multiple LLMs, benchmarks, and uncertainty metrics.

Our key contributions are outlined as follows:\\
\textbf{1. We introduce DebUnc}, a framework designed to quantify and communicate the uncertainty of LLM agents within multi-agent debates.  \\
\textbf{2. We propose a novel attention-scaling mechanism} that guides the model's focus towards more confident agents, serving as an alternative to conveying uncertainty through textual prompts. \\
\textbf{3. We evaluate DebUnc across multiple LLMs, benchmarks, and uncertainty metrics}, and find that debates using attention-scaling to communicate confidence outperform those using textual prompts to communicate confidence, as well as debates without any uncertainty communication.  \\
\textbf{4. We provide insights into future improvements}, exploring how more robust uncertainty metrics could further enhance debate performance.

\section{Related Work}

LLMs are known for their overconfidence and their tendency to provide responses to any user query, regardless of their certainty. This often leads to the generation of factual inaccuracies, known as hallucinations, where the information provided by the model is incorrect or unsupported by the data on which it was trained \citep{trust_feelings, to_believe_not_to_believe, llms_know_hallucinations, hallucination_adverserial, aichberger2024semantically}. Ensuring factual accuracy is crucial for building trust in LLM-based systems and expanding their use in real-world applications. As a result, there has been a surge in research focused understanding the mechanisms behind hallucinations and developing strategies to mitigate them \citep{mitigation_hallucinations_selfReflection, distillation_reducing_hallucinations, mitigation_hallucination_instruction_tuning}.

\subsection{Uncertainty in LLMs}\label{sec:relwunc}
Some current research efforts to mitigate hallucinations focus on measuring a model's uncertainty and enhancing its self-awareness \citep{llm_mostly_know, knowledge-of-knowledge, selfaware}.  Accurately measuring a model’s confidence would provide users with clearer guidance on when to trust its output \citep{teaching_unc_words, sayself} and enable language agents to better determine when to access external resources \citep{uala}. 

However, accurately measuring model uncertainty remains an open challenge, with ongoing research exploring new uncertainty metrics. We identify three primary approaches to this problem:

\textbf{Token Probability-Based Uncertainty Metrics}assess uncertainty by examining the probabilities assigned to each token generated by the model. High token probabilities (near 1) indicate strong model confidence, while lower probabilities suggest uncertainty, as the model spreads probability across multiple tokens. Notable methods in this category include Mean Token Entropy, Perplexity \citep{entropyperplexity}, SAR \citep{sar}, RDE \citep{rde}, and Claim-Conditioned Probability \citep{claimed_condition_prob}.

\definecolor{googreen}{RGB}{182, 215, 168}
\sethlcolor{googreen}
\begin{figure*}[h]
    \centering
    \includegraphics[width=\textwidth]{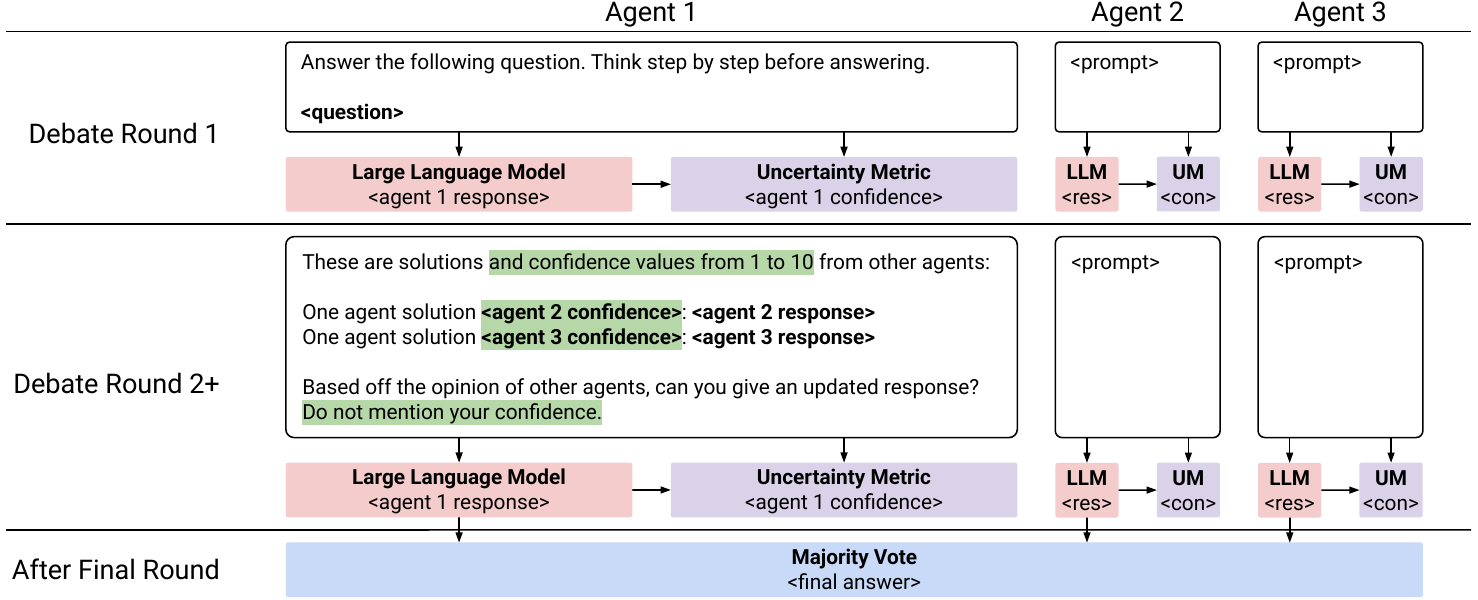}
    \caption{Illustration of the modified multi-agent debate involving three agents. In the first round, each agent independently generates a response to the question, which is evaluated for confidence using an uncertainty metric. The prompt for following rounds includes the responses from other agents in the previous round. Sections of the prompt highlighted in \hl{green} are used only with the \nameref{confinprompt} method. Each agent retains access to its complete chat history throughout the debate. After the final round, a majority vote determines the final answer.}
    \label{fig:pipeline}
\end{figure*}

\textbf{LLM-Generated Uncertainty Metrics} involve the model explicitly expressing its uncertainty in its response. \citet{teaching_unc_words} explored this by fine-tuning GPT-3 \citep{gpt3} to provide both an answer and a corresponding confidence level. Alternatively, the model can be prompted to express its uncertainty without explicit training. \citet{justaskforcalibration} found that this approach can can outperform token-probability-based methods when applied to LLMs fine-tuned with reinforcement learning from human feedback \citep{rlhf}. However, \citet{canllmexpressunc} report lower performance compared to token probability-based methods on GPT-3.

\textbf{Sampling-Based Uncertainty Metrics} evaluate uncertainty by generating multiple outputs and analyzing the distribution of meanings across them. Consistency in meaning across samples indicates high confidence, while variations suggest lower confidence. These methods generally outperform token probability-based metrics, but are more resource-intensive due to the need for multiple generations. Examples include Semantic Entropy \citep{semantic_entropy}, LUQ \citep{luq}, and other metrics that evaluate meaning diversity \citep{numsets}.

\subsection{Multi-Agent Debate}
With the increased accessibility of LLMs and improvements in their inference times, LLMs are being integrated into more complex systems as autonomous agents \citep{autogen, camel, metaGPT}. A critical component of these agent-based systems is the collaboration mechanism, where models engage in debate with one another. These mechanisms are currently being studied \citep{collaboration_exploration_llm} and have been shown to foster more divergent thinking \citep{divergent_thinking}, enhance reasoning and factual accuracy \citep{multiagentdebate}, and lead to more reliable evaluations \citep{debate_evaluators}. Through discussions, the LLMs can refine their outputs, ultimately achieving higher levels of agreement and producing more factually accurate text \citep{hallucination_debate, dont_hallucinate_abstain}.

\cite{ciphers} recognized that text is not be the most effective communication mechanism for LLM agents, since information is lost during the token sampling process, and demonstrated how LLMs can communicate through embeddings. ReConcile \cite{reconcile} explored the integration of agent confidence in multi-agent debates, relying on LLM agents to self-report their confidence, which was communicated to other agents through prompts. Building on these ideas, we employ uncertainty metrics to estimate agent confidence and explore both prompting and attention scaling to convey this confidence.

\section{Method}
In multi-agent LLM debates, agents often produce confident-sounding yet inaccurate responses, potentially misleading other agents and leading to consensus on incorrect answers \citep{multiagentdebate}. Our goal is to guide agents on which opinions to prioritize based on uncertainty levels. Our modified debate pipeline, illustrated in \autoref{fig:pipeline}, operates as follows: in each round of debate, every agent generates a response and its uncertainty is assessed. In the subsequent round, each agent shares its response and uncertainty with all other agents. We evaluate on three uncertainty metrics and three methods for communicating agent uncertainty.

\subsection{Uncertainty Metrics}
Uncertainty metrics assess an LLM's confidence in its responses: high uncertainty indicates low confidence and potential unreliability, while low uncertainty suggests greater confidence and reliability. These metrics generally fall into one of three categories: token probability-based, sampling-based, or LLM-generated methods. For more details, see \nameref{sec:relwunc}. In our experiments, we focus on token probability-based metrics due to their efficiency and simplicity, as they require only a single generation and do not rely on the model's ability to express uncertainty, unlike sampling-based and LLM-generated metrics, respectively. However, our methods could be used with any uncertainty metric.

Specifically, we selected Mean Token Entropy \citep{entropyperplexity} for its simplicity and TokenSAR \citep{sar}, a more advanced approach that recognizes that certain tokens contribute more significantly to a sequence's meaning than others. We utilized implementations from LM-Polygraph, a framework that provides implementations for various uncertainty metrics \citep{lmpolygraph}. Lastly, to evaluate the potential of future uncertainty metrics, we include a third "Oracle" uncertainty metric in our analysis.

\paragraph{Mean Token Entropy.}
One of the simplest and most computationally efficient uncertainty metrics is Mean Token Entropy\citep{entropyperplexity}. It is calculated as the average entropy across all tokens generated, with the entropy $H$ of a single token $X$ defined as:
\[
H(X) = -\sum_{x \in V} p(x) \log p(x)
\]
Here, $V$ denotes the model's vocabulary. Entropy is maximized when $p(x)$ is uniform over all tokens in the vocabulary, indicating maximum uncertainty. It is minimized when one token has a probability of 1 and all other tokens have a probability of 0, indicating complete certainty in the selected token.

\paragraph{TokenSAR.}
\cite{sar} recognized that some tokens contribute more to a text's meaning than others and introduced TokenSAR, an uncertainty metric that accounts for this. TokenSAR is defined as the weighted average of the negative log probabilities of each generated token, with weights determined by the relevance scores of the tokens.
\[
\text{TokenSAR} = \sum_i^N -\log p(t_i)R(t_i)
\]
Here, $N$ is the number of tokens generated, $t_i$ is the $i$-th token, and $R(t_i)$ is the relevance of token $t_i$. To compute each token's relevance, RoBERTa-large \citep{roberta} must be run $N$ times in total. This is more computationally expensive than calculating mean token entropy, but significantly less costly than metrics requiring multiple generations.

\paragraph{Oracle.}
While the uncertainty metrics discussed provide valuable insights into agent uncertainty, they are not flawless, and future advancements will likely lead to more accurate metrics. To assess the potential effectiveness of our uncertainty communication methods with improved uncertainty metrics, we include an "Oracle" metric in our experiments, which simulates an ideal uncertainty metric. This metric yields low uncertainty when the agent is correct and high uncertainty when the agent is incorrect, and is defined as follows:
\begin{figure*}[h]
    \centering
    \includegraphics[width=\textwidth]{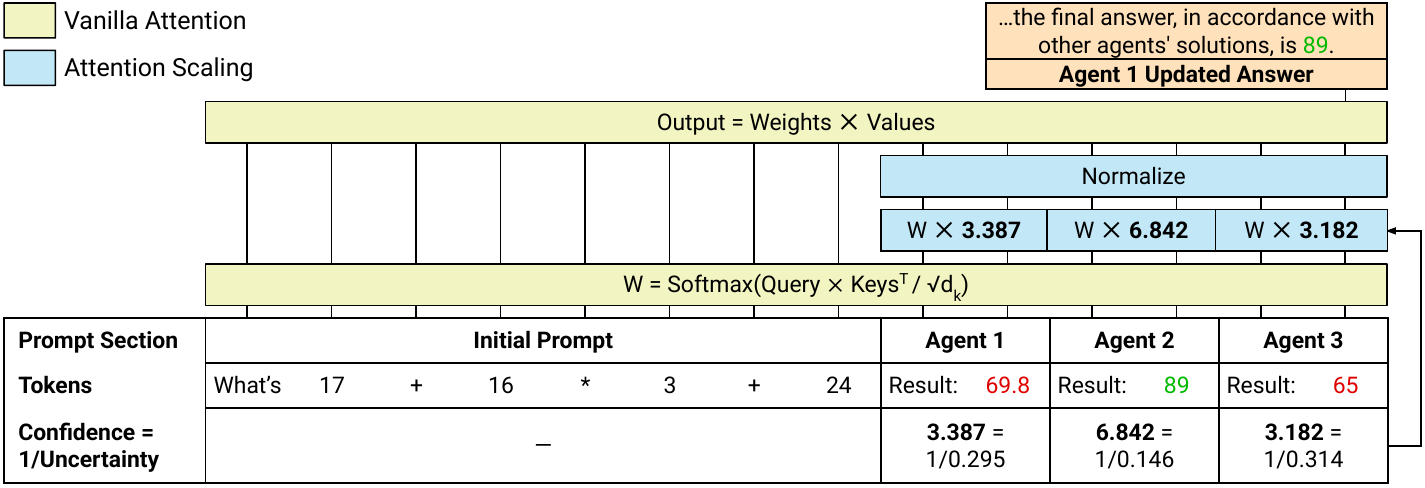}
    \caption{Illustration of the \protect\hyperlink{attnall}{Attention-All} method from the perspective of Agent 1. As the second debate round begins, the model's context includes the initial prompt and each agent's responses. Agent 2 provided a correct response with lower uncertainty than Agents 1 and 3, who responded incorrectly. Because Agent 2 had a lower uncertainty, the attention weights for tokens constituting Agent 2’s response will be increased, while those for tokens from Agent 1 and Agent 3's responses will be decreased. This led Agent 1 to switch to the correct answer.}
    \label{fig:attn}
\end{figure*}
\[
\text{uncertainty} =
\begin{cases}
0 & \text{if the response is correct} \\
\infty & \text{if the response is incorrect}
\end{cases}
\]
In practice, using 0 and $\infty$ could cause issues with our uncertainty communication methods, so we detail exactly how this metric is applied in the following subsection. It is also important to note that this metric requires knowledge of the ground truth answer, making it impractical for real-world use. Instead, it serves to evaluate the effectiveness of our uncertainty communication methods independently of the performance of the uncertainty metrics themselves, and allows us to anticipate how improvements in uncertainty metrics could affect debate performance.

\subsection{Uncertainty Communication}\label{sec:incorpunc}
After computing the uncertainty of each agent, we need to communicate it to the other agents. We explore several methods to incorporate these uncertainties into the following debate round.

\paragraph{Confidence in Prompt.}\label{confinprompt}
One approach is to include the uncertainties directly in the text prompt for the next debate round, as shown in \autoref{fig:pipeline}. However, it may be more natural to express confidence, where higher is better, rather than uncertainty, where lower is better. Humans, for example, often describe their confidence on a scale from 1 to 10. Since LLMs are trained on human data, they may exhibit a similar preference.

Mean Token Entropy and TokenSAR both yield non-negative uncertainties, but their scales differ: Mean Token Entropy has an upper bound that depends on the model's vocabulary size, while TokenSAR has an unbounded maximum. Because of this, absolute uncertainty values are less informative than the relative differences in uncertainty between agents.

In order to convert uncertainty values into confidence values, we perform the following steps. Given a list of uncertainties $u$ for $n$ agents, where $u_i$ is the uncertainty of agent $i$, we first invert them to obtain unscaled confidence values $c_i = \frac{1}{u_i}$. We then scale these values such that the average confidence $s_i$ of all agents is 5:

% \begin{align*}
\[s_i = \frac{c_i}{\sum_{j=1}^{n} r_j} \cdot (5n-1) + \frac{1}{n} \]
% \end{align*}

\setlength{\tabcolsep}{5pt} % Default is usually 6pt
\begin{table*}
\centering
% \small
% \begingroup
% \setlength{\tabcolsep}{5pt} % Reduce column spacing only for this table
\begin{tabular}{llcccccc}
\toprule
\textbf{Metric} & \textbf{Method} & \textbf{MMLU-0} & \textbf{MMLU-5} & \textbf{GSM8k} & \textbf{Truthful} & \textbf{Arithmetic} & \textbf{Average}\\
\midrule
---       & Standard     & 0.52 \small{$\pm$ 0.02} & 0.54 \small{$\pm$ 0.02} & 0.51 \small{$\pm$ 0.05} & 0.60 \small{$\pm$ 0.03} & 0.48 \small{$\pm$ 0.03} & 0.53 \small{$\pm$ 0.01} \\
\midrule
Entropy   & Prompt       & 0.52 \small{$\pm$ 0.02} & 0.55 \small{$\pm$ 0.03} & \textbf{0.54} \small{$\pm$ 0.04} & 0.60 \small{$\pm$ 0.04} & 0.48 \small{$\pm$ 0.05} & 0.54 \small{$\pm$ 0.01} \\
          & Attn-Others  & \textbf{0.54} \small{$\pm$ 0.02} & 0.57 \small{$\pm$ 0.04} & 0.49 \small{$\pm$ 0.03} & \textbf{0.61} \small{$\pm$ 0.03} & \textbf{0.52} \small{$\pm$ 0.09} & 0.54 \small{$\pm$ 0.02} \\
          & Attn-All     & 0.53 \small{$\pm$ 0.05} & \textbf{0.57} \small{$\pm$ 0.04} & \textbf{0.54} \small{$\pm$ 0.03} & 0.60 \small{$\pm$ 0.03} & \textbf{0.52} \small{$\pm$ 0.10} & \textbf{0.55} \small{$\pm$ 0.02} \\
\midrule
SAR       & Prompt       & \textbf{0.54} \small{$\pm$ 0.03} & 0.55 \small{$\pm$ 0.05} & 0.50 \small{$\pm$ 0.03} & 0.62 \small{$\pm$ 0.01} & 0.46 \small{$\pm$ 0.04} & 0.53 \small{$\pm$ 0.01} \\
          & Attn-Others  & 0.53 \small{$\pm$ 0.04} & \textbf{0.56} \small{$\pm$ 0.04} & 0.50 \small{$\pm$ 0.03} & \textbf{0.63} \small{$\pm$ 0.03} & 0.50 \small{$\pm$ 0.06} & 0.54 \small{$\pm$ 0.01} \\
          & Attn-All     & 0.53 \small{$\pm$ 0.02} & 0.55 \small{$\pm$ 0.04} & \textbf{0.53} \small{$\pm$ 0.06} & 0.61 \small{$\pm$ 0.03} & \textbf{0.50} \small{$\pm$ 0.10} & \textbf{0.54} \small{$\pm$ 0.02} \\
\midrule
Oracle*   & Prompt       & 0.56 \small{$\pm$ 0.03} & 0.56 \small{$\pm$ 0.03} & 0.55 \small{$\pm$ 0.04} & 0.63 \small{$\pm$ 0.03} & 0.54 \small{$\pm$ 0.06} & 0.57 \small{$\pm$ 0.01} \\
          & Attn-Others  & 0.61 \small{$\pm$ 0.04} & 0.67 \small{$\pm$ 0.03} & 0.64 \small{$\pm$ 0.04} & 0.63 \small{$\pm$ 0.03} & 0.65 \small{$\pm$ 0.04} & 0.64 \small{$\pm$ 0.01} \\
          & Attn-All     & \textbf{0.62} \small{$\pm$ 0.03} & \textbf{0.68} \small{$\pm$ 0.06} & \textbf{0.66} \small{$\pm$ 0.05} & \textbf{0.65} \small{$\pm$ 0.02} & \textbf{0.73} \small{$\pm$ 0.05} & \textbf{0.67} \small{$\pm$ 0.01} \\
\bottomrule
\end{tabular}
\caption{Accuracy comparison $\pm$ 95\% CI across various benchmarks using different uncertainty metrics (* the oracle metric simulates a perfect metric) and communication methods with Mistral-7B. `MMLU-0' and `MMLU-5' represent 0 and 5-shot performance on MMLU respectively. The other benchmarks used zero-shot prompting. The `Average' column shows the average performance over all benchmarks and the \textcolor{green}{\% increase} over the standard debate.}
\label{table:comparison}
% \endgroup
\end{table*}
Finally, we clamp the confidence levels to the range of 1 to 10 and round to the nearest integer. When using the Oracle uncertainty metric, we set the confidence to 1 if the agent was incorrect and to 10 if the agent was correct.

\paragraph{Attention Scaling.}\label{attnscm}

Another approach to communicate confidence levels is to modify the LLM's token generation process to account for each agent's confidence. Many LLMs use Transformer decoder layers that generate an embedding for the last token and use this embedding to predict the next token \citep{gpt1}. This embedding is determined by the attention mechanism, which creates "query," "key," and "value" vectors for each token.

The similarity between the "query" vector of the last token and the "key" vector of each token is used to compute a weight for every token. These weights are normalized with a softmax function to ensure they sum to 1, and are used to create the output embedding, which is the weighted sum of the value vectors of each token \citep{attention}. The weight of each token determines its influence on the next token generated. By modifying these weights, we can adjust the model's focus to each token in the input. In multi-agent debates, this allows us to shift the model's focus towards the responses from more confident agents. 

After each debate round, we have responses from each agent. In the next round, each agent's prompt will include these responses. We also compute the uncertainty of each agent using an uncertainty metric. In the next round, as the LLM generates its response, it computes the normalized attention weights for each preceding token. We divide the weight $w_i$ of every token from agent $j$ by $m_j$, which is defined as the inverse of agent $j$'s uncertainty when using Mean Token Entropy or TokenSAR. When using the Oracle metric, to avoid divide-by-zero errors, we set the $m_j$ to $10^{-5}$ if agent $j$ was incorrect and $1$ if agent $j$ was correct.

Formally, the scaled weight $a_i$ for token i is:
\[
a_i =
\begin{cases} 
w_i \cdot m_j & \text{if } i \in t_j, \text{ for any agent } j \\
w_i & \text{otherwise}
\end{cases}
\]

Here, $t_j$ is the set of token indices from agent $j$. After this, the attention weights may no longer sum to 1, so another normalization step is needed. We only normalize the weights at indices that were scaled, leaving the weights of other tokens unchanged. The final weight $f_i$ for token $i$ is:
\[
f_i =
\begin{cases}
    a_i \cdot \frac{\sum_{j=1}^{n} \sum_{k \in t_j}^{} w_k}{\sum_{j=1}^{n} \sum_{k \in t_j}^{} a_k} & \parbox[c]{2.5cm}{$\text{if } i \in t_j, \\ \text{for any agent } j$} \\
    a_i & \text{otherwise}
\end{cases}
\]

For simplicity, we apply attention scaling to only the responses from the previous round. For instance, in a three-round debate, attention would be rescaled for the responses from the first round during the second round, and for the responses from the second round during the third round. In the third round of debate, attention would not be rescaled to the first-round responses.

Additionally, in order to prevent divide-by-zero errors during normalization, attention is not scaled when computing the embeddings for tokens within the prompt; it is only scaled when generating answer tokens.

We explore two variants of attention scaling:
\begin{itemize}
    \item \textbf{Attention-Others}, where agent $i$ only rescales attention to other agents' tokens $t_j \mid j \neq i$
    \item \hypertarget{attnall}{\textbf{Attention-All}}, where agent $i$ rescales attention to other agents and itself, illustrated in \autoref{fig:attn}
\end{itemize}

\section{Experiment Design}\label{experiment_design}

To evaluate these methods, an open source LLM is required, as implementing the attention scaling requires modifications to the model source code. In addition, the uncertainty metrics used are based on token probabilities that may not be readily available from closed-source models.

We ran all of the experiments on Mistral-7B-Instruct-v0.2 \citep{mistral}, and re-evaluated the most insightful ones on Llama-3-8B-Instruct \citep{llama3}. Tokens were sampled with a temperature of 1 to ensure responses varied. Instead of evaluating on the full datasets, which would be too expensive, we sampled 100 questions from each. We evaluated every combination of uncertainty metric and communication method on these samples five times, and report the mean accuracy across the five runs $\pm$ the 95\% confidence interval.
\begin{table}[!ht]
\centering
\label{tab:experiment_design}
\begin{tabular}{ll}
\toprule
\textbf{Parameter}      & \textbf{Details}                                                   \\ \midrule
\textbf{LLMs}           & Mistral-7B, Llama-3-8B                                            \\
\textbf{Temperature}    & 1                             \\
\textbf{Unc. Metrics}   & Entropy, TokenSAR, Oracle \\
\textbf{Benchmarks}     & \textbf{MMLU} Broad knowledge \\
                        & \textbf{GSM8k} Math problems \\
                        & \textbf{TruthfulQA} Misconceptions \\
                        & \textbf{Arithmetic} Random equations                           \\
\textbf{Prompting}      & MMLU: 0/5-shot for Mistral                                   \\
                        & Others: 0-shot                                        \\
\textbf{Questions}      & 100 sampled per benchmark                                         \\
\textbf{Repetitions}    & 5 runs per method                                                \\
\textbf{Debate Setup}   & 3 agents, 3 rounds, same LLM                                      \\ \bottomrule
\end{tabular}
\caption{Experiment Design Summary}
\end{table}

\begin{table*}[h]
    \centering
    \begin{tabular}{llccccc}
        \toprule
        \textbf{Metric} & \textbf{Method}  & \textbf{MMLU-0} & \textbf{GSM8k} & \textbf{TruthfulQA} & \textbf{Arithmetic} & \textbf{Average} \\
\midrule
---       & Standard     & 0.65 \small{$\pm$ 0.03} & 0.81 \small{$\pm$ 0.04} & 0.52 \small{$\pm$ 0.02} & 0.52 \small{$\pm$ 0.05} & 0.63 \small{$\pm$ 0.01} \\
\midrule
Entropy   & Prompt       & 0.61 \small{$\pm$ 0.02} & \textbf{0.84} \small{$\pm$ 0.05} & 0.54 \small{$\pm$ 0.05} & 0.53 \small{$\pm$ 0.05} & 0.63 \small{$\pm$ 0.02} \\
          & Attn-Others  & 0.64 \small{$\pm$ 0.06} & 0.81 \small{$\pm$ 0.03} & \textbf{0.56} \small{$\pm$ 0.05} & 0.53 \small{$\pm$ 0.08} & 0.63 \small{$\pm$ 0.02} \\
          & Attn-All     & \textbf{0.66} \small{$\pm$ 0.03} & 0.81 \small{$\pm$ 0.03} & 0.56 \small{$\pm$ 0.04} & \textbf{0.53} \small{$\pm$ 0.05} & \textbf{0.64} \small{$\pm$ 0.01} \\
\midrule
Oracle*   & Prompt       & 0.67 \small{$\pm$ 0.03} & 0.87 \small{$\pm$ 0.02} & 0.58 \small{$\pm$ 0.04} & 0.55 \small{$\pm$ 0.05} & 0.67 \small{$\pm$ 0.01} \\
          & Attn-Others  & \textbf{0.78} \small{$\pm$ 0.05} & 0.90 \small{$\pm$ 0.03} & 0.67 \small{$\pm$ 0.02} & 0.56 \small{$\pm$ 0.06} & \textbf{0.73} \small{$\pm$ 0.02} \\
          & Attn-All     & 0.75 \small{$\pm$ 0.03} & \textbf{0.90} \small{$\pm$ 0.02} & \textbf{0.68} \small{$\pm$ 0.04} & \textbf{0.56} \small{$\pm$ 0.05} & 0.72 \small{$\pm$ 0.01} \\
        \bottomrule
    \end{tabular}
    \caption{Accuracy comparison $\pm$ 95\% CI across various benchmarks using different uncertainty metrics (* the oracle metric simulates a perfect metric) and communication methods with Llama-3-8B. Zero-shot prompting was used for all benchmarks. The `Average' column shows the average performance over all benchmarks and the \textcolor{green}{\% increase} over the standard debate.}
    \label{table:llama}
\end{table*}

We evaluated on the following benchmarks:
\begin{enumerate}
\item \textbf{MMLU} \citep{mmlu}: A dataset of multiple-choice questions across various subjects.
\item \textbf{GSM8k} \citep{gsm8k}: A dataset of free-response grade school math problems.
\item \textbf{TruthfulQA} \citep{truthfulqa}: A multiple-choice dataset testing the model's susceptibility to common misconceptions.
\item \textbf{Arithmetic}: An randomly generated set of arithmetic problems in the form $a + b \cdot c + d$ for the Mistral model and $a + b \cdot c + d - e \cdot f$ for the Llama model, where $0 \leq a, b, c, d < 30$.
\end{enumerate}

For MMLU on Mistral-7B, we tested both zero-shot and 5-shot prompting with examples from the original MMLU repository's\footnote{\url{https://github.com/hendrycks/test}} dev set. For other benchmarks, we used only zero-shot prompting. The prompts used are shown in \autoref{appendix:prompts}.

The debates involved three agents and three rounds, with all agents using the same LLM. In the first round, each agent answered the question individually. In the following rounds, each agent was given other agents' responses, and used this information to refine their answers. A full example debate is shown in \autoref{appendix:debate}. The final answer was determined by a majority vote after the third round. This workflow is illustrated in \autoref{fig:pipeline}.

\begin{figure*}[h]
    \centering
    \includegraphics[width=\textwidth]{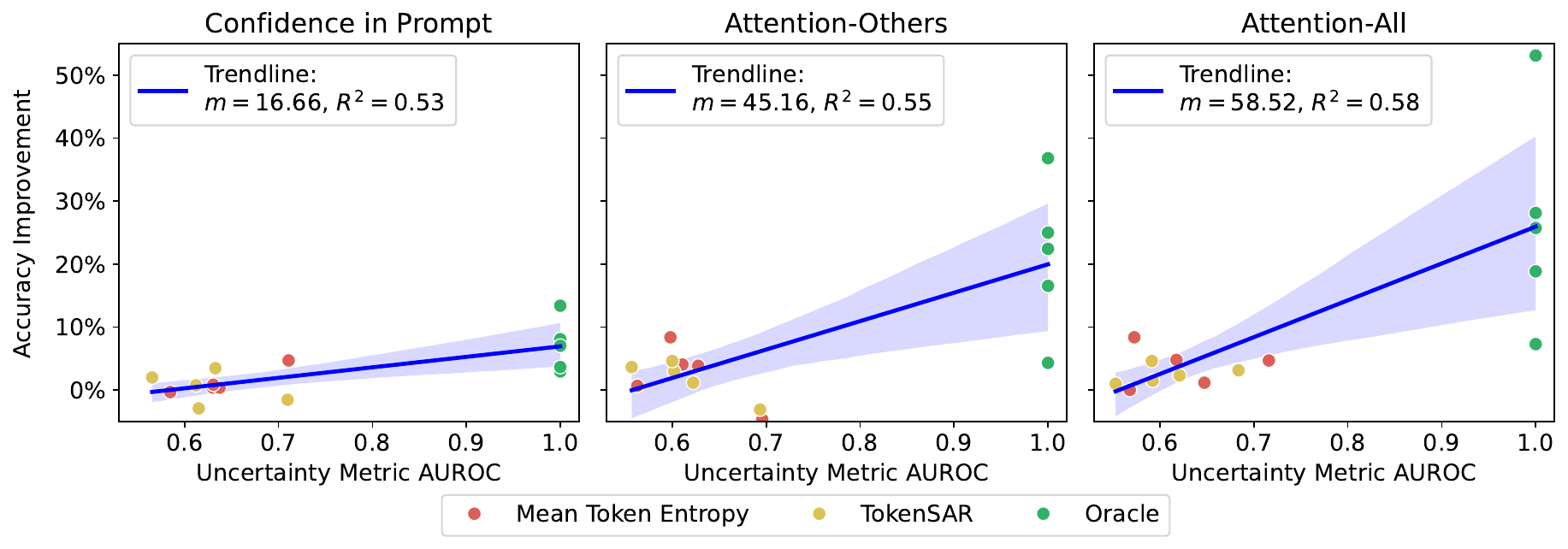}
    \caption{Plots showing the percent increase in accuracy over standard debate versus uncertainty metric AUROC for a given combination of benchmark, uncertainty metric, and trial using Mistral-7B. A higher AUROC indicates better metric performance. The plots are titled by uncertainty incorporation method and color-coded by the uncertainty metric used. The trendlines show that attention-based methods, especially Attention-All, lead to more substantial performance gains as AUROC increases compared to methods that incorporate confidence directly into the prompt.}
    \label{fig:uncvacc}
\end{figure*}

\begin{figure*}
    \centering
    \includegraphics[width=\textwidth]{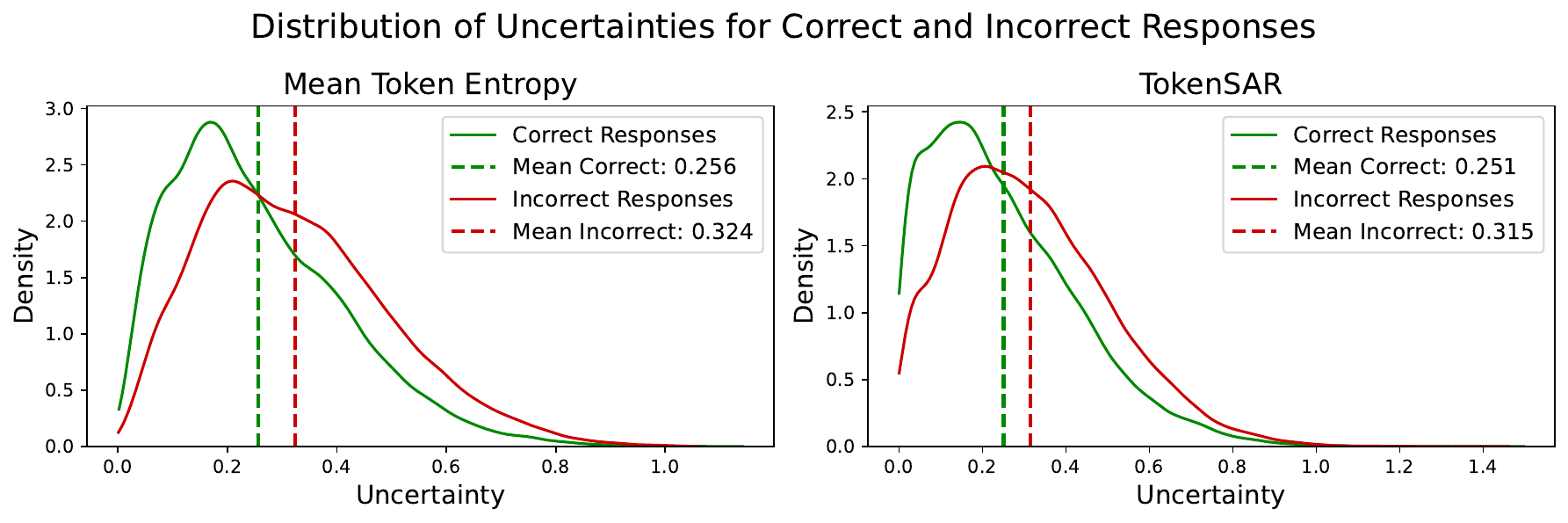}
    \caption{Distribution of uncertainties for correct and incorrect answers across all Mistral-7B experiments, as measured by the uncertainty metrics Mean Token Entropy and TokenSAR. Generally, correct answers exhibit lower uncertainties than incorrect ones, indicating that although not perfect, uncertainty metrics are useful for distinguishing between accurate responses and those where the agent may be hallucinating.}
    \label{fig:uncdist}
\end{figure*}

\section{Results}
In this section, we first analyze the effectiveness of each uncertainty incorporation method, and then analyze the effectiveness of the uncertainty metrics.

\subsection{Uncertainty Incorporation Methods}\label{res:uncincmeth}
\autoref{table:comparison} presents a comparison of the results obtained with Mistral-7B using different combinations of uncertainty metrics (Mean Token Entropy, TokenSAR, and Oracle) and methods (Confidence in Prompt, Attention-Others, and Attention-All). As a baseline, the performance of a standard 3-agent, 3-round debate without any uncertainty metrics is also shown. Overall, Attention-All was the top-performing method, achieving the highest average accuracy across all three uncertainty metrics. It was the only method that consistently matched or exceeded the performance of the standard multi-agent debate on all benchmarks. \autoref{table:llama}  presents the results using Llama-3-8B. The findings show that with the Oracle uncertainty metric, the attention scaling methods significantly outperformed confidence communication via prompting. However, when using mean token entropy, the performance gains were minimal, suggesting that mean token entropy may not be as effective on Llama-3-8B as it was on Mistral-7B.

As shown in \autoref{fig:uncvacc}, Attention-All demonstrates the most significant accuracy improvements as the AUROC of the uncertainty metric increases, with a slope of 0.59 compared to 0.45 for Attention-Others and 0.17 for Confidence in Prompt. The accuracy improvement ratio compares the method's accuracy to the accuracy observed in a standard debate. AUROC, the area under the receiver operating characteristic curve, represents the probability that a correct answer is assigned a lower uncertainty than an incorrect one. A random uncertainty metric would have an AUROC of 0.5, while a perfect one would have an AUROC of 1.

\subsection{Uncertainty Metrics}
The best-performing uncertainty metric was, by definition, the Oracle metric. Mean Token Entropy ranked next, with debates using it typically achieving higher average accuracies than debates using TokenSAR, as shown in \autoref{table:comparison}. Mean Token Entropy achieved an average AUROC across all experiments of 0.627, compared to 0.617 for TokenSAR. Full AUROC results are shown in \autoref{appendix:unc_metrics}.

To further analyze the uncertainty metrics and understand their impact, \autoref{fig:uncdist} presents the distributions of uncertainties for responses containing correct answers versus incorrect answers from Mistral-7B across all benchmarks and uncertainty incorporation methods.

The left plot illustrates the uncertainties when using Mean Token Entropy, while the right plot depicts uncertainties when using TokenSAR. As shown in \autoref{table:comparison}, Mean Token Entropy achieved slightly higher accuracy than TokenSAR on average. Examining the average uncertainties for correct and incorrect responses, Mean Token Entropy has a ratio of average correct to average incorrect of 1.264, compared to 1.258 for TokenSAR. This suggests that Mean Token Entropy slightly outperforms TokenSAR in differentiating between correct and incorrect responses, while also being less computationally demanding. However, the performance difference is minimal and falls within the margin of error.

\section{Conclusion}
We introduce DebUnc, a framework that integrates model uncertainty into multi-agent debates. By incorporating confidence levels, agents can assess the reliability of their peers and make more informed decisions when responses differ. Confidence is estimated using an uncertainty metric and conveyed through either attention scaling, which adjusts token weights based on confidence levels, or direct inclusion in prompts. Our results show that attention scaling outperforms textual prompts, highlighting that LLMs can process information beyond traditional text inputs. This technique is broadly applicable to scenarios where certain parts of the input should be prioritized, and we encourage further exploration of its potential.

\section{Limitations}
One limitation of DebUnc is its reliance on open-source LLMs, as attention scaling requires code modifications, and token probabilities may not be accessible from proprietary models. Additionally, attention scaling is sensitive to the order of agent responses within the prompt. For example, if agent 2's response precedes agent 3's, unidirectional attention allows agent 2 to influence agent 3's tokens, potentially leaking information from agent 2 even if agent 3 is more confident.

\section*{Ethics Statement}

Deploying large language models (LLMs) to the general public requires them to be trustworthy and reliable, particularly in high-risk scenarios. This concern gains more importance when LLMs interact within agent frameworks. We foresee a future where LLMs interact with other LLMs to achieve complex tasks. In this work, we take a step towards achieving reliable communication among agents by incorporating uncertainty estimations.

% Bibliography entries for the entire Anthology, followed by custom entries
%\bibliography{anthology,custom}
% Custom bibliography entries only
\bibliography{custom}

\onecolumn
\appendix

\section{Example Debate}\label{appendix:debate}
This is a full debate on zero-shot MMLU using Mistral-7B as the LLM. It uses the Confidence in Prompt method for incorporating uncertainty with the Oracle uncertainty metric. For conciseness, the user prompts for rounds 2 and 3 use ``<agent $i$ response>" to reference the response from agent $i$ in the previous round.

\begin{xltabular}{\textwidth}{>{\raggedright\arraybackslash\sfns}X >{\raggedright\arraybackslash\sfns}X >{\raggedright\arraybackslash\sfns}X}

\toprule
\multicolumn{1}{c}{\textbf{Agent 1 (cont.)}} & \multicolumn{1}{c}{\textbf{Agent 2 (cont.)}} & \multicolumn{1}{c}{\textbf{Agent 3 (cont.)}} \\
\midrule
\endhead

\toprule
\multicolumn{1}{c}{\textbf{Agent 1}} & \multicolumn{1}{c}{\textbf{Agent 2}} & \multicolumn{1}{c}{\textbf{Agent 3}} \\
\midrule
\endfirsthead

\multicolumn{3}{c}{\textit{User, Round 1}} \\
\midrule
\multicolumn{3}{p{\textwidth}}{\sfns \AgentPromptRoundOne} \\
\midrule
\multicolumn{3}{c}{\textit{Assistant, Round 1}} \\
\midrule
\AgentOneResponseRoundOne & \AgentTwoResponseRoundOne & \AgentThreeResponseRoundOne \\
\midrule 
\end{xltabular}

\pagebreak
\begin{xltabular}{\textwidth}{>{\raggedright\arraybackslash\sfns}X >{\raggedright\arraybackslash\sfns}X >{\raggedright\arraybackslash\sfns}X}

\toprule
\multicolumn{1}{c}{\textbf{Agent 1 (cont.)}} & \multicolumn{1}{c}{\textbf{Agent 2 (cont.)}} & \multicolumn{1}{c}{\textbf{Agent 3 (cont.)}} \\
\midrule
\endhead

\multicolumn{3}{c}{\textit{User, Round 2}} \\
\midrule
\AgentOnePromptRoundTwo & \AgentTwoPromptRoundTwo & \AgentThreePromptRoundTwo \\
\midrule
\multicolumn{3}{c}{\textit{Assistant, Round 2}} \\
\midrule
\AgentOneResponseRoundTwo & \AgentTwoResponseRoundTwo & \AgentThreeResponseRoundTwo \\
\midrule
\end{xltabular}

\pagebreak
\begin{xltabular}{\textwidth}{>{\raggedright\arraybackslash\sfns}X >{\raggedright\arraybackslash\sfns}X >{\raggedright\arraybackslash\sfns}X}

\toprule
\multicolumn{1}{c}{\textbf{Agent 1 (cont.)}} & \multicolumn{1}{c}{\textbf{Agent 2 (cont.)}} & \multicolumn{1}{c}{\textbf{Agent 3 (cont.)}} \\
\midrule
\endhead

\multicolumn{3}{c}{\textit{User, Round 3}} \\
\midrule
\AgentOnePromptRoundThree & \AgentTwoPromptRoundThree & \AgentThreePromptRoundThree \\
\midrule
\multicolumn{3}{c}{\textit{Assistant, Round 3}} \\
\midrule
\AgentOneResponseRoundThree & \AgentTwoResponseRoundThree & \AgentThreeResponseRoundThree \\
\bottomrule
\end{xltabular}

\newpage

\section{Full Uncertainty Metric Results}\label{appendix:unc_metrics}
\begin{table*}[h]
\centering
\begin{tabular}{llcccccc}
\toprule
\textbf{Estimator} & \textbf{Method} & \textbf{MMLU-0} & \textbf{MMLU-5} & \textbf{GSM8k} & \textbf{Arithmetic} & \textbf{TruthfulQA} & \textbf{Avg.}\\
\midrule
Entropy & Prompt & 0.630 & 0.637 & 0.711 & 0.631 & 0.585 & 0.639 \\
& Attn-Others& 0.628 & 0.611 & 0.696 & 0.598 & 0.563 & 0.619 \\
& Attn-All& 0.647 & 0.618 & 0.716 & 0.573 & 0.568 & 0.624 \\
\midrule
TokenSAR & Prompt & 0.633 & 0.612 & 0.710 & 0.615 & 0.566 & 0.627 \\
& Attn-Others& 0.622 & 0.602 & 0.694 & 0.600 & 0.557 & 0.615 \\
& Attn-All& 0.621 & 0.592 & 0.684 & 0.591 & 0.553 & 0.608 \\
\midrule
Oracle & Prompt & 1.000 & 1.000 & 1.000 & 1.000 & 1.000 & 1.000 \\
& Attn-Others& 1.000 & 1.000 & 1.000 & 1.000 & 1.000 & 1.000 \\
& Attn-All& 1.000 & 1.000 & 1.000 & 1.000 & 1.000 & 1.000 \\
\bottomrule
\end{tabular}
\caption{This table displays the uncertainty metric AUROC values for each experiment run with Mistral-7B. `MMLU-0' denotes zero-shot prompting on MMLU, while `MMLU-5' represents 5-shot prompting. The other benchmarks used zero-shot prompting. The `Avg.' column shows the average AUROC for all tests. By definition, the Oracle metric achieved perfect AUROC scores. Among the remaining metrics, Mean Token Entropy delivered slightly better performance than TokenSAR despite being cheaper to run.}
\end{table*}

\newpage

\section{Prompts}\label{appendix:prompts}
The table below displays the prompts used to facilitate the debates. Text in \textcolor{green}{green} is only included when using the Confidence in Prompt method, and text in \textcolor{blue}{blue} is specific to each debate.

\begin{xltabular}{\textwidth}{ll>{\small}X}
\toprule
\textbf{Benchmark} & \textbf{Prompt Type} & \textbf{\normalsize Prompt} \\
\midrule
\endhead
MMLU & Initial, 0-shot & Answer the following multiple choice question. The last line of your response should be of the following format: 'Answer: \$LETTER' (without quotes) where LETTER is one of ABCD. Think step by step before answering. \newline \newline \textcolor{blue}{<question>} \\
\cmidrule{2-3}
& Initial, 5-shot & Answer the following multiple choice question.\newline \newline Examples:\newline \newline \textcolor{blue}{<5 example questions and answers>} \newline \newline ---\newline \newline YOUR TASK\newline \newline Answer the following question. Think step by step before answering. The last line of your response should be of the following format: 'Answer: \$LETTER' (without quotes) where LETTER is one of ABCD.\newline \newline \textcolor{blue}{<question>} \\
\cmidrule{2-3}
& Debate & These are solutions \textcolor{green}{and confidence values from 1 to 10 (higher means more confident)} to the problem from other agents: \newline \newline  One agent solution \textcolor{green}{(confidence level is $c$)}: \textcolor{blue}{<agent response>}\newline \newline  One agent solution \textcolor{green}{(confidence level is $c$)}: \textcolor{blue}{<agent response>}\newline \newline Based off the opinion of other agents, can you give an updated response? \textcolor{green}{Do not mention your confidence.} Think step by step before answering. The last line of your response should be of the following format: 'Answer: \$LETTER' (without quotes) where LETTER is one of ABCD. \\
\midrule
GSM8k & Initial & Answer the following math problem. The last line of your response should be of the following format: 'Answer: \$INTEGER' (without quotes) where INTEGER is the integer answer. Think step by step before answering.\newline \newline \textcolor{blue}{<question>} \\
\cmidrule{2-3}
& Debate & These are solutions \textcolor{green}{and confidence values from 1 to 10 (higher means more confident)} to the problem from other agents: \newline \newline  One agent solution \textcolor{green}{(confidence level is $c$)}: \textcolor{blue}{<agent response>}\newline \newline  One agent solution \textcolor{green}{(confidence level is $c$)}: \textcolor{blue}{<agent response>}\newline \newline Based off the opinion of other agents, can you provide an updated response? The original problem is:\newline \newline \textcolor{blue}{<question>}\newline \newline \textcolor{green}{Do not mention your confidence.} The last line of your response should be of the following format: 'Answer: \$INTEGER' (without quotes) where INTEGER is the integer answer. \\
\midrule
\pagebreak
Arithmetic & Initial & What is the result of \textcolor{blue}{<a+b*c+d>}? State the final answer at the end of your response. \\
\cmidrule{2-3}
& Debate & These are solutions \textcolor{green}{and confidence values from 1 to 10 (higher means more confident)} to the problem from other agents: \newline \newline  One agent solution \textcolor{green}{(confidence level is $c$)}: \textcolor{blue}{<agent response>}\newline \newline  One agent solution \textcolor{green}{(confidence level is $c$)}: \textcolor{blue}{<agent response>}\newline \newline Based off the opinion of other agents, can you provide an updated answer? \textcolor{green}{Do not mention your confidence.} State the final answer at the end of your response. \\
\midrule
TruthfulQA & Initial & Answer the following multiple choice question:\newline \newline \textcolor{blue}{<question>}\newline \newline \newline Think step by step before answering. The last line of your response should be of the following format: 'Answer: \$LETTER' (without quotes) where LETTER is one of \textcolor{blue}{<options>}. \\
\cmidrule{2-3}
& Debate & These are the selections \textcolor{green}{and confidence values from 1 to 10 (higher means more confident)} from other agents: \newline \newline  One agent solution \textcolor{green}{(confidence level is $c$)}: \textcolor{blue}{<agent response>}\newline \newline  One agent solution \textcolor{green}{(confidence level is $c$)}: \textcolor{blue}{<agent response>}\newline \newline Can you double check that your response is correct? \textcolor{green}{Do not mention your confidence.} The last line of your response should be of the following format: 'Answer: \$LETTER' (without quotes) where LETTER is one of \textcolor{blue}{<options>}. \\
\bottomrule
\end{xltabular}

\end{document}